\documentclass[lettersize,journal]{IEEEtran}
\usepackage{amsmath,amsfonts}
\usepackage{spconf}
\usepackage{bm}
\usepackage{algorithmic}
\usepackage{algorithm}
\usepackage{array}
\usepackage[caption=false,font=normalsize,labelfont=sf,textfont=sf]{subfig}
\usepackage{textcomp}
\usepackage{stfloats}
\usepackage{url}
\usepackage{verbatim}
\usepackage{cite}
\usepackage{mathrsfs}
\usepackage{booktabs}
\usepackage{multirow}
\usepackage{graphicx}
\usepackage{cite}
\usepackage{hyperref}
\usepackage{orcidlink}
\definecolor{darkgreen}{rgb}{0.0, 0.5, 0.0}
\usepackage[table,xcdraw]{xcolor}

\title{Bi-Modal Textual Prompt Learning for Vision-Language Models in Remote Sensing}
\name{Pankhi Kashyap$^{\star}$, Mainak Singha$^{\dagger}$, Biplab Banerjee$^{\star}$}
\address{
$^{\star}$Indian Institute of Technology Bombay, 
$^{\dagger}$University of Trento
}

\begin{document}

\maketitle

\begin{abstract}
Prompt learning (PL) has emerged as an effective strategy to adapt vision-language models (VLMs), such as CLIP, for downstream tasks under limited supervision. While PL has demonstrated strong generalization on natural image datasets, its transferability to remote sensing (RS) imagery remains underexplored. RS data present unique challenges, including multi-label scenes, high intra-class variability, and diverse spatial resolutions, that hinder the direct applicability of existing PL methods. In particular, current prompt-based approaches often struggle to identify dominant semantic cues and fail to generalize to novel classes in RS scenarios. To address these challenges, we propose BiMoRS, a lightweight bi-modal prompt learning framework tailored for RS tasks. BiMoRS employs a frozen image captioning model (e.g., BLIP-2) to extract textual semantic summaries from RS images. These captions are tokenized using a BERT tokenizer and fused with high-level visual features from the CLIP encoder. A lightweight cross-attention module then conditions a learnable query prompt on the fused textual-visual representation, yielding contextualized prompts without altering the CLIP backbone. We evaluate BiMoRS on four RS datasets across three domain generalization (DG) tasks and observe consistent performance gains, outperforming strong baselines by up to 2\% on average. 
Codes are available at \url{https://github.com/ipankhi/BiMoRS}.
\end{abstract}


\section{Introduction}
Remote sensing imagery serves as a critical resource for applications such as environmental monitoring, urban planning, agriculture, and disaster response \cite{zhu2017deep, li2022deep}, while providing rich spatial and semantic information at varying resolutions. However, deep learning models trained on a specific RS domain often fail to generalize to unseen domains with variations in geography, sensor modality, illumination, seasonality, and spatial granularity, obscuring fine-grained RS scene classification.

Recent advances in \textit{vision-language models (VLMs)} such as CLIP\cite{clip} show strong generalization in natural images through contrastive pre-training of aligned image-text representations. Their ability to support zero-shot and few-shot learning makes them attractive for data-scarce RS settings, but direct transfer is suboptimal due to the \textit{top-down perspective}, \textit{multi-label nature}, and \textit{dense spatial semantics} unique to RS. For example, a single RS image may contain rivers, highways, vegetation, and built structures, rendering class-level text prompts inadequate and prone to semantic mismatch \cite{tuia2016domain, zhao2020review, chen2020cross, gondara2016domain}.

Adaptations like RemoteCLIP\cite{remoteclip} and RSCLIP\cite{rsclip} fine-tune VLMs on large RS-caption datasets, but incur high costs and limited scalability, while still lacking image-specific adaptability at inference. \textit{Prompt learning} offers a lightweight alternative. Methods like CoOp\cite{coop} and CoCoOp\cite{cocoop} steer frozen VLMs using learnable prompts, but these are static or class-conditioned, limiting power in multicategory RS scenes. Extensions such as APPLeNet\cite{applenet} employ attention for RS prompt tuning, yet struggle with semantic ambiguity and spatial intricacy. Importantly, RS imagery requires both spatially adaptive and semantically diverse prompting strategies \cite{csaw} to fully capture its fine-grained complexity. Despite progress, two key gaps remain:
(i) \textit{Lack of image-specific prompt adaptation}, as the methods overlook the rich, context-dependent semantics in RS imagery.
(ii) \textit{Dependence on domain-specific pretraining}, requiring expensive RS-caption datasets and retraining, which limits scalability and broader applicability.

\noindent \textbf{Our Proposal.} To overcome these limitations, we propose BiMoRS, a \textit{Bi-Modal Prompt Learning} framework designed specifically for RS image classification under domain shifts. Unlike static class-level prompt methods, BiMoRS generates \textit{image-adaptive prompts} by leveraging both textual and visual semantics derived directly from the input image.
BiMoRS operates by:
(i)  Using a \textit{frozen image captioning model} (e.g., BLIP-2\cite{blip2}) to extract natural language descriptions of RS scenes, which encapsulate contextual elements absent from class labels.
(ii) Tokenizing these captions with a \textit{BERT tokenizer} \cite{bert}, and projecting them into the CLIP embedding space via a \textit{learnable projector}.
(iii) Concurrently extracting \textit{high-level visual features} from the frozen CLIP encoder and projecting them similarly.
(iv) Applying a \textit{cross-attention mechanism} where a learnable query prompt attends over the fused textual and visual tokens, producing dynamic and contextualized prompts without modifying the CLIP or BLIP-2 backbones. The bi-modal design of BiMoRS allows to adjust its prompt representation based on the specific content of each image, resulting in improved contextual grounding and generalization, without sacrificing computational efficiency. We demonstrate strong generalization performance across multiple RS datasets and settings, achieving up to 2\% improvement in accuracy while using 80\% fewer learnable parameters compared to recent state-of-the-art approaches.

\begin{figure*}[htbp]
    \centering
    \includegraphics[width=\textwidth, trim=19 0 19 0, clip]{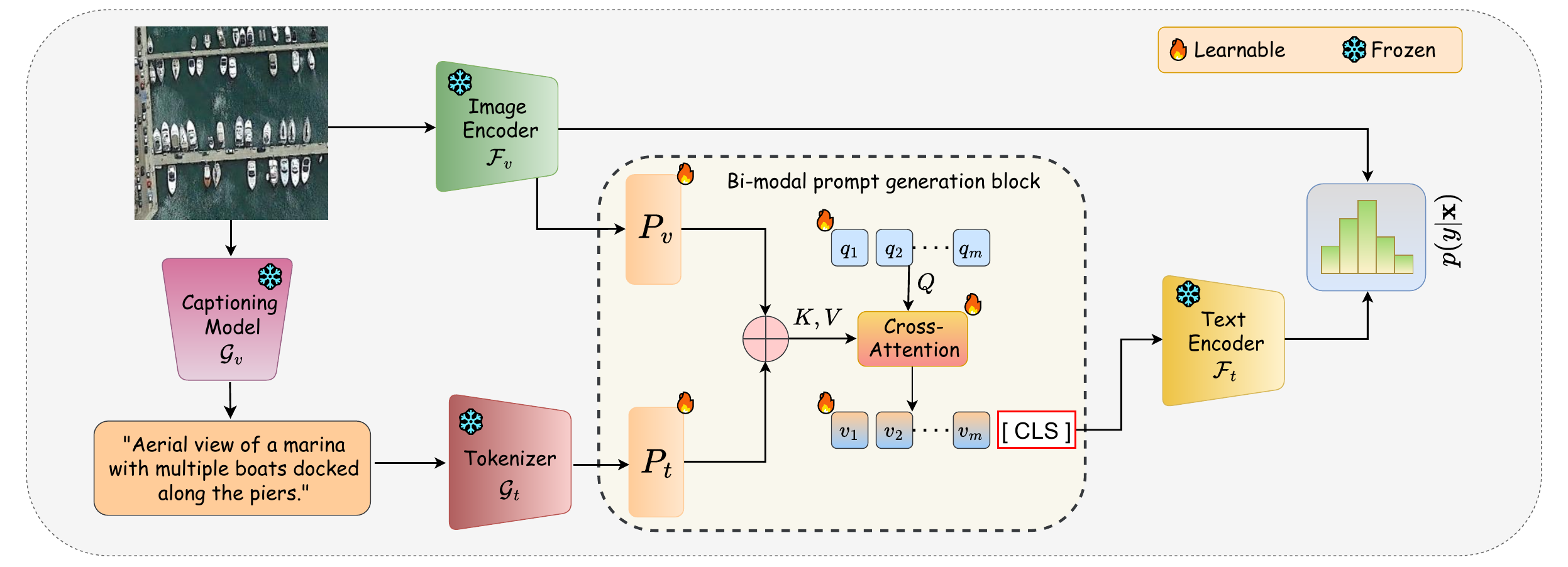} 
    \caption{\textbf{Model architecture of BiMoRS}, comprises of two main components: (i) firstly, we propose to capture the visual information through an image-captioning model $\mathcal{G}_v$ as attribute descriptions and tokenize them using a tokenizer $\mathcal{G}_t$, and (ii) secondly, we propose a lightweight cross-attention module $\bm{\mathrm{CA}}$, that effectively combines both of the visual features from the final layer of $\mathcal{F}_v$ and tokenized descriptions, rescaled through projectors $P_v$ and $P_t$ respectively. Finally, $\bm{\mathrm{CA}}$ generates bi-modal text prompts that efficaciously generalize BiMoRS through textual prompting with the better alignment of V-L representations.}
    \label{fig:main_fig}
    \vspace{-0.5cm}
\end{figure*}

\vspace{-0.30cm}
\section{Proposed Methodology}

\subsection{Problem definition}

We study two generalization settings: (i) Base-to-new class generalization: source and target label sets are disjoint ($\mathcal{C}_{\text{Seen}} \cap \mathcal{C}_{\text{Unseen}} = \emptyset$), where $\mathcal{C}_{\text{Seen}} \cup \mathcal{C}_{\text{Unseen}} = \mathcal{C}$ i.e. total number of classes. The model must recognize unseen classes at inference using only images and class names (zero-shot). (ii) Domain generalization (DG): label spaces are shared ($\mathcal{C}_{\text{source}} = \mathcal{C}_{\text{target}}$), but distributions differ ($P(\mathcal{D}_t) \neq P(\mathcal{D}_s^i)$). The aim is to learn domain-invariant features. For DG, we evaluate on two cases: cross-dataset generalization (CD) and single-source multi-target (SSMT).

\vspace{-0.3cm}
\subsection{Going through the BiMoRS framework}
BiMoRS extends CLIP with bi-modal, image-conditioned prompts that fuse textual and visual cues from RS imagery to improve contextual understanding while remaining lightweight and modular. Its key components are: (i) frozen image captioning model $\mathcal{G}_v$ that extracts high-level semantic descriptions, (ii) a pretrained tokenizer $\mathcal{G}_t$ that encodes captions into contextual embeddings, (iii) two lightweight projection heads ($P_t$, $P_v$) that align text and vision features in a shared space, and (iv) a cross-attention module $\bm{\mathrm{CA}}$ that fuses modalities to generate adaptive prompts, focusing on the most informative features for the task.

\noindent \textbf{Bi-modal Textual Prompt Learning.} Given an input image $x$, BiMoRS first generates a natural language caption using a frozen image captioning model $\mathcal{G}_v$. This semantic description is denoted as $\mathcal{A} = \mathcal{G}_v(x)$. The caption $\mathcal{A}$ is then tokenized using a pretrained language tokenizer $\mathcal{G}_t$, and subsequently projected into a shared textual embedding space via a lightweight projection head $P_t$. The resulting textual embedding is given by $\mathcal{A}' = P_t(\mathcal{G}_t(\mathcal{A})) \in \mathbb{R}^{1 \times d}$,
where $d$ represents the dimensionality of the embedding space. In parallel, BiMoRS extracts visual features from the penultimate layer (i.e., $(L-1)$-th layer) of the frozen CLIP image encoder, denoted as $\mathcal{F}_v^{L-1}(x)$. These features are also projected into the same embedding space using a visual projection head $P_v$; $\mathcal{V}' = P_v(\mathcal{F}_v^{L-1}(x)) \in \mathbb{R}^{1 \times d}$. 

To enable bi-modal interaction, the textual and visual embeddings are concatenated to form a unified representation, defined as $\mathcal{B}' = [\mathcal{A}'; \mathcal{V}'] \in \mathbb{R}^{2 \times d}$. Meanwhile, a learnable set of query prompt tokens $Q' \in \mathbb{R}^{m \times d}$ is initialized, where $m$ is the number of prompt tokens. These prompts are adaptively refined through a multi-head cross-attention module $\bm{\mathrm{CA}}$, which enables the query tokens to attend to the bi-modal context $\mathcal{B}'$. The output of the cross-attention mechanism is defined as,
\begin{align}
    \mathcal{Q} = \bm{\mathrm{CA}}(Q', \mathcal{B}') = \bm{\mathrm{MHA}}(Q', K_{\mathcal{B}'}, V_{\mathcal{B}'}) + Q'
    \label{eq:cross_attn}
\end{align}
\vspace{-0.30cm}
where $\bm{\mathrm{MHA}}$ denotes multi-head attention. The key and value projections of the bi-modal tokens are computed as
\begin{align}
    K_{\mathcal{B}'} &= \mathcal{B}' W_K, \quad V_{\mathcal{B}'} = \mathcal{B}' W_V \\
    \bm{\mathrm{Att}}(Q', K, V) &= \text{softmax}\left( \frac{Q' W_Q K^\top}{\sqrt{d}} \right) V
    \label{eq:attn}
\end{align}

where $W_Q, W_K, W_V \in \mathbb{R}^{d \times d}$ are learnable projection matrices corresponding to the query, key, and value transformations. The cross-attention mechanism empowers the prompt tokens to capture task-relevant cues by dynamically integrating visual and linguistic information in a unified manner.

The resulting prompts $\mathcal{Q} = [v_1, \dots, v_m]$ encode image-specific contextual semantics. These $m$ tokens replace the static context tokens in the standard CLIP prompt template (e.g., ``a photo of a [CLS]''), following the approach of CoOp \cite{coop}. The final prompts are then fed into CLIP’s frozen text encoder $\mathcal{F}_t$ to obtain class-level textual representations i.e., $\mathcal{F}_t(\mathcal{Q}) \in \mathbb{R}^{C \times d}$.

\noindent \textbf{Training objective and inference.} We train only the parameters of $P_t$, $P_v$, and $\bm{\mathrm{CA}}$, while keeping the CLIP backbone ($\mathcal{F}_v$, $\mathcal{F}_t$) and the captioning pipeline ($\mathcal{G}_v$, $\mathcal{G}_t$) completely frozen. Let $\mathcal{Q}_y$ denote the final prompt embedding for class $y$. Given a test image $x$, its classification score is computed via cosine similarity between the visual embedding and each class prompt,
\begin{equation}
    p(y|x) = \frac{ \exp(\text{sim}(\mathcal{F}_v(x), \mathcal{F}_t(\mathcal{Q}_y)) / \tau) }{ \sum_{k=1}^C \exp(\text{sim}(\mathcal{F}_v(x), \mathcal{F}_t(\mathcal{Q}_k)) / \tau) }
    \label{eq:prob}
\end{equation}

where $\text{sim}(\cdot,\cdot)$ denotes cosine similarity and $\tau$ is a temperature parameter. The model is optimized using cross-entropy loss over the source dataset $\mathcal{D}_s$,
\begin{equation}
    \mathcal{L} = - \mathbb{E}_{(x, y) \sim \mathcal{D}_s} \log p(y|x)
    \label{eq:loss}
\end{equation}

During inference, the predicted class label $\hat{y}_t$ for a test sample $x_t$ is given by, $\hat{y}_t = \arg\max_{y \in \mathcal{C}} p(y | x_t)$.

\noindent The BiMoRS framework thus enables efficient, adaptive, and content-aware prompt tuning by leveraging both visual and textual modalities, significantly improving generalization performance in remote sensing tasks without retraining the backbone models.

\vspace{-0.30cm}
\section{Experimental Evaluation}
\noindent \textbf{Experimental Protocol.}
BiMoRS contains three learnable components: a cross-attention module ($\bm{\mathrm{CA}}$) with $h = 4$ heads and a two-layer feedforward block (LayerNorm-ReLU-Linear), and two projection heads ($P_v$ and $P_t$) that map visual and textual features from 768 to 512 dimensions. $P_v$ uses global average pooling for CLIP features, while $P_t$ is a single linear layer for tokenized captions. We use CLIP ViT-B/16 as the frozen backbone, BLIP-2 for captioning, and a BERT tokenizer for text encoding, both frozen during training. BiMoRS is trained for 10 epochs with SGD (learning rate $2 \times 10^{-4}$, warm-up $1 \times 10^{-5}$), batch size 4, and 16 shots per class; prompt tokens ($m=4$) are initialized as \texttt{``a photo of a [CLS]''}. Results are averaged over three seeds (top-1 accuracy). We evaluate in three settings: (i) \textit{Base-to-New (B2N)} i.e. testing on novel classes; (ii) \textit{Cross-Dataset (CD)} i.e. training/testing on different datasets; (iii) \textit{Single-Source Multi-Target (SSMT)} i.e. training on one dataset, evaluating on others with shared classes to isolate domain shift. We evaluate all of the settings on the given RS datasets similar to \cite{applenet}. 

\noindent \textbf{Base-to-New (B2N) generalization.} Table~\ref{tab:results_basenew} presents the performance on the B2N task, measured using the harmonic mean (H) between the inference results of base and novel classes. BiMoRS consistently delivers the best performance across all datasets, surpassing zero-shot CLIP by 24.50\% on base classes and 5.12\% on novel classes. Compared to the baselines, BiMoRS outperforms them by at least of ~1\% in terms of the average harmonic mean.
Notably, it outperforms APPLeNet by 10.01\% while having only 20\% of its trainable parameters. Against more complex, layer-wise adaptation methods like MaPLe and TCP, BiMoRS achieves comparable or better performance, improving new class accuracy by up to 2\%, while preserving architectural simplicity. Remarkably, with just 1M trainable parameters, BiMoRS achieves the highest overall accuracy among all compared methods. In contrast, models such as APPLeNet and MaPLe rely on 3-5 times more parameters yet offer marginal accuracy gains, while lighter models like TCP still fall slightly short in overall performance.
\begin{table*}[htbp]
\centering
\scriptsize
\caption{Comparison of BiMoRS with state-of-the-art methods on the base-to-new (B2N) generalization task. Accuracy is reported for Base, New classes, and Harmonic mean (H). $\Delta$ shows the difference from the second-best method.}
\scalebox{1}{
\begin{tabular}{lc ccc ccc ccc ccc ccc}
\toprule
\multirow{2}{*}{\textbf{Method}} & \multicolumn{1}{c}{{\textbf{Trainable}}} & \multicolumn{3}{c}{\textbf{PatternNet}} & \multicolumn{3}{c}{\textbf{RSICD}} & \multicolumn{3}{c}{\textbf{RESISC45}} & \multicolumn{3}{c}{\textbf{MLRSNet}} & \multicolumn{3}{c}{\textbf{Average}} \\
 & \multicolumn{1}{c}{\textbf{Params}} & \textbf{Base} & \textbf{New} & \textbf{H} & \textbf{Base} & \textbf{New} & \textbf{H} & \textbf{Base} & \textbf{New} & \textbf{H} & \textbf{Base} & \textbf{New} & \textbf{H} & \textbf{Base} & \textbf{New} & \textbf{H} \\
\midrule
CLIP\cite{clip}      &   0     & 73.30 & \textbf{67.70} & 70.38 & 60.70 & 65.20 & 62.86 & 70.20 & 61.50 & 65.56 & 65.10 & 55.60 & 59.97  & 67.32  & 62.50 & 64.69 \\
CoOp\cite{coop}      &   2048     & 95.59 & 49.93 & 65.59 & 95.67 & 57.43 & 71.77 & 89.83 & 61.41 & 72.94 & 82.96 & 54.49 & 65.77 & 91.01 & 55.81 & 69.01 \\
CoCoOp\cite{cocoop}  &  0.035M      & 94.80 & 55.90 & 70.32 & 95.50 & 63.40 & 76.20 & 90.50 & 61.40 & 73.16 & 82.10 & 59.45 & 68.96 & 90.72 & 60.03 & 72.16\\
ProGrad\cite{prograd}&   8192     & 90.00 & 63.95 & 74.77 & 94.27 & 66.56 & 78.02 & 88.14 & 70.93 & 78.60 & 79.74 & 58.39 & 67.41 & 88.03 & 64.95 & 74.70 \\
APPLeNet\cite{applenet} &  5M   & 96.50 & 50.30 & 66.13 & 96.20 & 60.80 & 74.50 & \textbf{91.00} & 55.10 & 68.63 & 86.40 & 48.50 & 62.12 & \textbf{92.52} & 53.67  & 67.84 \\
MaPLe\cite{maple}    &   3M     & \textbf{97.50} & 57.80 & 72.57 & 94.90 & 69.20 & 80.03 & 90.70 & 68.10 & 77.79 & 86.40 & 55.60 & 67.65 & 92.37 & 62.67 & 74.51\\
TCP\cite{tcp}        &  0.3M      & 96.20 & 64.50 & 77.22 & 96.39 & 69.80 & 80.97 & 90.30 & 68.50 & 77.90 & \textbf{86.90} & 60.80 & 71.54 & 92.44  & 65.90 & 76.90 \\
\midrule
\rowcolor{gray!20} \textbf{BiMoRS} & 1M & 96.10 & 66.60 & \textbf{78.67} & \textbf{96.40} & \textbf{70.80} & \textbf{81.64} & 90.80 & \textbf{69.80} & \textbf{78.92} & 84.00 & \textbf{63.30} & \textbf{72.19} & 91.82 & \textbf{67.62}  & \textbf{77.85} \\
\rowcolor{gray!20} $\Delta$ & & \textcolor{red}{-1.40} & \textcolor{red}{-1.10} & \textcolor{darkgreen}{+1.45}
                           & \textcolor{darkgreen}{+0.01} & \textcolor{darkgreen}{+1.00} & \textcolor{darkgreen}{+0.67}
                     
                           & \textcolor{darkgreen}{+0.50} & \textcolor{darkgreen}{+1.30} & \textcolor{darkgreen}{+1.02}
                           & \textcolor{red}{-2.90} & \textcolor{darkgreen}{+2.50} & \textcolor{darkgreen}{+0.65} & \textcolor{red}{-0.70}   & \textcolor{darkgreen}{+1.72} & \textcolor{darkgreen}{+0.95} \\
\bottomrule
\end{tabular}}
\vspace{-0.2cm}
\label{tab:results_basenew}
\end{table*}

\begin{figure*}[htbp]
    \centering
    \includegraphics[width=\textwidth]{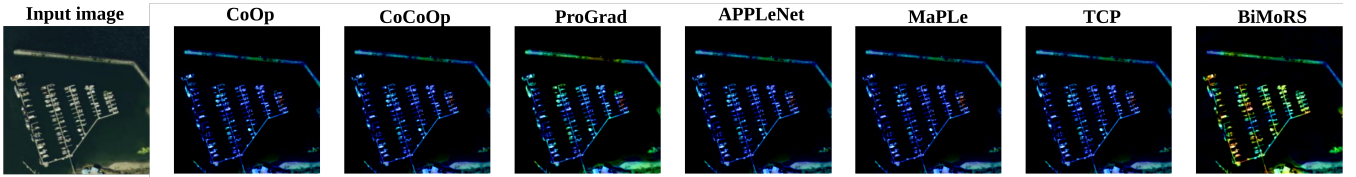}
    \vspace{-0.9cm}
    \caption{Comparison of attention maps generated by SOTA methods on a sample of class `Harbor' using Grad-CAM \cite{gradcam}. }
    \vspace{-0.7cm}
    \label{fig:attn_maps}
\end{figure*}

\noindent \textbf{Cross-Dataset (CD) generalization.} 
As shown in Table~\ref{tab:results_cross}, BiMoRS demonstrates the strongest generalization across all source-target dataset pairs. When trained on a single dataset, BiMoRS achieves notable improvements in average target accuracy over APPLeNet by 9.38\% across all datasets. Although deep prompt tuning methods such as TCP and MaPLe exhibit strong performance, BiMoRS remains competitive or outperforms them in most scenarios, underscoring the efficacy of our multimodal, input-level prompting strategy.

\begin{table}[!ht]
\centering
\scriptsize{
    \centering
    \caption{Comparison of BiMoRS with SOTA methods on the cross-dataset generalization task. The average results for the target datasets are reported under their corresponding source datasets. The best results are highlighted in \textbf{bold}.}
    \vspace{0.2cm}
    \scalebox{0.8}{
    \begin{tabular}{lccccc} 
\toprule                               
\textbf{Method} &
\textbf{PatternNet} & 
\textbf{RSICD}   & 
\textbf{RESISC45}   & 
\textbf{MLRSNet} &
\textbf{Average} \\ 
\midrule
CLIP\cite{clip}                              & 54.10                  & 54.93                          & 49.50                            &51.46 &52.49   \\  
CoOp\cite{coop}                               & 53.06               & 56.12                       & 64.62                         & 63.29 &59.27   \\  
CoCoOp\cite{cocoop}                            & 54.73               & 52.96                       & 64.90                         & 62.66 & 58.81 \\  
ProGrad\cite{prograd}                            & 53.32               & 55.23                       & 64.00                         & 61.26 & 58.45   \\  
APPLeNet\cite{applenet}                           & 45.83               & 51.13                       & 63.55                         & 56.53  & 54.26  \\  
MaPLe\cite{maple}                              & 58.73               & 59.73                       & 65.98                         & 64.03  & 62.11  \\  
TCP\cite{tcp}                                & 59.66               & 60.95                       & 66.03                         & 64.55  & 62.79  \\ 
\midrule
\textbf{\cellcolor{gray!20} BiMoRS}  & \cellcolor{gray!20} \textbf{60.13}      & \cellcolor{gray!20}\textbf{61.66}              & \cellcolor{gray!20}\textbf{66.86}                & \cellcolor{gray!20}\textbf{65.91} & \cellcolor{gray!20}\textbf{63.64} \\        \cellcolor{gray!20}$\Delta$ & 
\cellcolor{gray!20}\textcolor{darkgreen}{+0.47} & 
\cellcolor{gray!20}\textcolor{darkgreen}{+0.71} & 
\cellcolor{gray!20}\textcolor{darkgreen}{+0.83} & 
\cellcolor{gray!20}\textcolor{darkgreen}{+1.36} &
\cellcolor{gray!20}\textcolor{darkgreen}{+0.85}\\ 
\bottomrule
\end{tabular}}
\vspace{-0.5cm}
\label{tab:results_cross}}
\end{table}

\noindent\textbf{Single-Source Multi-Target (SSMT) generalization.}
In the SSMT setting, BiMoRS once again demonstrates consistent superiority, as shown in Table~\ref{tab:results_domain}. When evaluated on 16 shared categories across unseen target domains, it achieves an average improvement of 0.6\% over the second-best method. This robustness highlights BiMoRS's ability to effectively bridge domain gaps using only lightweight prompt tuning, without any fine-tuning of the image encoder or backbone. The qualitative advantage is further illustrated in the attention map visualizations (Figure~\ref{fig:attn_maps}), where BiMoRS exhibits more precise focus on class-relevant regions compared to other baselines.

\begin{table}[]
\centering
\scriptsize{
    \centering
    \caption{Comparison of BiMoRS with state-of-the-art methods on the single-source domain generalization task. The average results for the target datasets are reported under their corresponding source datasets. The best results are highlighted in \textbf{bold}.} 
    \vspace{0.2cm}
    \scalebox{0.8}{
    \begin{tabular}{lccccc} 
\toprule                               
{\textbf{Method}} & 
\textbf{PatternNetv2} & 
\textbf{RSICDv2} & 
\textbf{RESISC45v2} &
\textbf{MLRSNetv2} &
\textbf{Average} \\ 
\midrule
CLIP\cite{clip}       &  73.02 & 74.82  & 73.02  & 73.62  & 73.62  \\  
CoOp\cite{coop}        & 81.88  & 82.70  & 80.80  & 87.65 & 83.25 \\  
CoCoOp\cite{cocoop}     & 78.70  & 83.63  & 87.73  & 90.50 & 85.14 \\  
ProGrad\cite{prograd}    & 82.98  & 83.43  & 86.24  & 89.27 & 85.48 \\  
APPLeNet\cite{applenet}                              & 81.76  & 81.60 & 86.73  & 88.76 & 84.71 \\  
MaPLe\cite{maple}     & 85.26  & 84.76 & 87.68  & 90.13 & 86.95\\
TCP\cite{tcp}       & 84.70  & 85.13 & 87.90  & 89.45 & 86.79\\
\midrule
\textbf{\cellcolor{gray!20} BiMoRS} & \cellcolor{gray!20}\textbf{85.46} & \cellcolor{gray!20}\textbf{85.91}  & \cellcolor{gray!20}\textbf{88.20}  & \cellcolor{gray!20}\textbf{90.66} & \cellcolor{gray!20}\textbf{87.55}  \\     
\cellcolor{gray!20}$\Delta$ & 
\cellcolor{gray!20}\textcolor{darkgreen}{+0.20} & 
\cellcolor{gray!20}\textcolor{darkgreen}{+0.78} & 
\cellcolor{gray!20}\textcolor{darkgreen}{+0.30} & 
\cellcolor{gray!20}\textcolor{darkgreen}{+0.16} &
\cellcolor{gray!20}\textcolor{darkgreen}{+0.60}\\ 
\bottomrule
\end{tabular}
}
\vspace{-0.8cm}
\label{tab:results_domain}}
\end{table}

\vspace{-0.4cm}
\section{Ablation analysis}
To dissect key architectural choices, we performed ablation studies examining both the impact of cross-attention in prompt generation and the effect of textual tokenization. \textit{\textbf{First}}, we evaluated four configurations of BiMoRS to understand the role of cross-attention: (i) removing both cross-attention and $P_t$, using only visual features to initialize prompts; (ii) using cross-attention with visual tokens as key-value pairs, so that prompt queries attend only to visual cues; (iii) using only tokenized captions as key-value inputs to isolate linguistic contributions; and (iv) the full BiMoRS setup, which fuses visual and textual features in the cross-attention module. As summarized in Table~\ref{tab:cross_attn}, jointly attending to both visual and textual cues consistently achieves the highest H-score, highlighting the power of multimodal fusion for generalization to novel categories. \textit{\textbf{Second}}, to probe the influence of tokenization, we compared BiMoRS variants using CLIP’s built-in tokenizer and BERT’s tokenizer (on BLIP-2 captions). While both tokenizers are offline and untrained, BERT’s captures richer linguistic structure and context, which is critical for remote sensing scenes with many co-occurring objects. As shown in Table~\ref{tab:text_enc}, BERT-tokenized prompts yield a clear performance boost especially for unseen classes, demonstrating that linguistically rich tokenization further enhances semantic alignment, even without additional fine-tuning.

\begin{table}[]
\centering
\scriptsize{
    \centering
    \caption{Comparison of different cross-attention configurations in BiMoRS on the PatternNet dataset.}
    \vspace{0.2cm}
    \scalebox{1.05}{
\begin{tabular}{cccc}
\toprule
& \multicolumn{3}{c}{\textbf{PatternNet}}\\ 
\cmidrule(l){2-4} 
\multirow{-2}{*}{\textbf{Method}} &\multicolumn{1}{c}{\textbf{Base}}&\multicolumn{1}{c}{\textbf{New}}&\multicolumn{1}{c}{\textbf{H}} \\ 
\midrule
w/o $\bm{\mathrm{CA}}$ and $P_t$        & \textbf{97.70} & 60.30 & 74.57 \\
$\bm{\mathrm{CA}}$ with $P_v$ only             & 96.50 & 51.60 & 67.24 \\
$\bm{\mathrm{CA}}$ with $P_t$ only             & 97.00 & 49.52 & 65.56 \\
\cellcolor{gray!20}$\bm{\mathrm{CA}}$ with $P_v$ \& $P_t$ & \cellcolor{gray!20}96.10 & \cellcolor{gray!20}\textbf{66.60} & \cellcolor{gray!20}\textbf{78.67} \\ 
\bottomrule
\end{tabular} }
\vspace{-0.6cm}
\label{tab:cross_attn}}
\end{table}

\vspace{-0.5cm}
\begin{table}[htbp]
\centering
\scriptsize{
    \centering
    \caption{Performance comparison on PatternNet with different caption token generation approaches in BiMoRS.}
    \vspace{0.2cm}
    \scalebox{1.05}{
\begin{tabular}{cccc}
\toprule
& \multicolumn{3}{c}{\textbf{PatternNet}}\\ 
\cmidrule(l){2-4} 
\multirow{-2}{*}{\textbf{Method}} &\multicolumn{1}{c}{\textbf{Base}}&\multicolumn{1}{c}{\textbf{New}}&\multicolumn{1}{c}{\textbf{H}} \\  

\midrule
CLIP text encoder         & 96.00 & 62.70 & 75.85 \\
\cellcolor{gray!20}Bert Tokenizer      & \cellcolor{gray!20}\textbf{96.10} & \cellcolor{gray!20}\textbf{66.60} & \cellcolor{gray!20}\textbf{78.67} \\ 
\bottomrule
\end{tabular} }
\label{tab:text_enc}}
\vspace{-0.5cm}
\end{table}

\section{Conclusions}
In this work, we present BiMoRS, a novel framework that introduces bi-modal, semantically rich prompts to enhance generalization tasks in remote sensing (RS). This method effectively aligns vision-language representations by extracting enhanced visual features from a pre-trained image-captioning model and training a lightweight cross-attention module for joint textual and visual semantics. We believe that our approach can be extended to other vision tasks such as segmentation and image retrieval for RS applications. \\
\noindent \textbf{Acknowledgement:} \textit{B. Banerjee acknowledges the support from ANRF grant no: CRG/2023/004389}

\bibliographystyle{IEEEtran}
\bibliography{refs}

\end{document}